%% file: main.tex
\documentclass[10pt,twocolumn,letterpaper]{article}
\usepackage{multirow}       % 用于 \multirow
\usepackage{booktabs}       % 用于三线表 (\toprule, \midrule, \bottomrule, \cmidrule)
\usepackage{graphicx}       % 用于 \resizebox
\usepackage{array}          % 用于复杂的列格式定义
\usepackage{placeins}

\usepackage{cvpr}

\definecolor{cvprblue}{rgb}{0.21,0.49,0.74}
\usepackage[pagebackref,breaklinks,colorlinks,allcolors=cvprblue]{hyperref}

\def\confName{CVPR}
\def\confYear{2027}

\title{NavTTT: Adaptive Parametric Memory via Hierarchical Test-Time Training for Vision-and-Language Navigation}

\author{%
{\normalsize
Xiaoyu Luo$^{1,2}$\thanks{Equal contribution.
This work was conducted during an internship at AGIBOT.} \quad
Chengxing Lin$^{2,3}$\footnotemark[1] \quad
Jiaxin Huang$^{2,4}$\thanks{Project leader.} \quad
Jiale Liu$^{5}$}\\
{\normalsize
Jingyu Guo$^{2,6}$ \quad
Sirui Chen$^{2,7}$ \quad
Aolin Li$^{2,8}$ \quad
Shaoli Huang$^{2}$\thanks{Corresponding author.}}\\
{\normalsize
Cheng-zhong Xu$^{1}$ \quad
Hui Kong$^{1}$}\\[0.5em]
{\small $^{1}$University of Macau \quad
        $^{2}$AGIBOT \quad
        $^{3}$UESTC \quad
        $^{4}$MBZUAI}\\
{\small $^{5}$University of Edinburgh \quad
        $^{6}$University of Melbourne}\\
{\small $^{7}$Nanyang Technological University \quad
        $^{8}$Wuhan University}\\[0.4em]
{\small \textbf{Project page:}\ 
\url{https://gamepeopleplay.github.io/NavTTT/}}
}

\begin{document}
\maketitle
\input{sec/0_abstract}    
\input{sec/1_intro}

\input{sec/2_related_work}
\input{sec/3_method}
\input{sec/4_experiment}
\input{sec/5_conclusion}
\clearpage
\clearpage % 再次清页，重置遗留的栏高

\raggedbottom
{
    \small
    \setlength{\bibsep}{1pt}
    \bibliographystyle{ieeenat_fullname}
    \bibliography{main}
}
\input{supplmental_matrerial}


\end{document}

%% file: sec/0_abstract.tex
\begin{abstract}
% Long-horizon vision-and-language navigation (VLN) requires agents
% to use visual evidence acquired earlier along a route, even after
% the corresponding frames leave the active visual context. We
% present NavTTT, which maintains an episode-local parametric memory
% through test-time training. Under a fixed write budget, NavTTT
% prioritizes task-relevant visual tokens and suppresses redundancy
% within each frame. It jointly optimizes memory representations
% of the retained visual evidence using semantic and spatial cues.
% Separate nonlinear fast weights at multiple decoder depths allow
% each augmented layer to encode and retrieve its own representation
% of the accumulated history. The decoder queries this memory to
% predict the current action chunk and commits the update from the
% current observation for the next decision. On the R2R-CE and RxR-CE
% Val-Unseen splits, NavTTT attains success rates of 66.7\% and
% 63.6\%, respectively, and improves navigation performance over
% static memory on both benchmarks. Real-robot trials illustrate its
% use in physical environments. These findings suggest that a
% fixed-capacity parametric memory can support long-horizon navigation
% by selectively retaining task-relevant evidence and making it
% accessible at multiple stages of action prediction.
Vision-and-language navigation (VLN) demands that agents accumulate and reason over streaming visual observations throughout long navigation episodes. Existing approaches struggle to maintain long-term history, constrained by the finite context capacity of transformer architectures. To solve the problem, fast-weight test-time training (TTT) provides a mechanism for encoding streaming observations into an evolving parameter state. However, existing TTT does not distinguish informative observations from repetitive visual content, allowing redundant tokens to interfere with the retention of information needed for subsequent decisions. We propose \textbf{Navigation-Conditioned Test-Time Training (NavTTT)}, which introduces adaptive memory formation and hierarchical memory integration for vision-and-language navigation. NavTTT leverages global navigation context to prioritize informative visual evidence and integrates language instructions with spatial information to construct spatially grounded memory representations. Episode-local nonlinear fast weights store these representations as Adaptive Parametric Memory (APM), preserving history beyond the visual context for later decisions. To reduce cross-layer coupling in memory formation, NavTTT maintains independent fast-weight memories across multiple decoder depths. This decoupled design prevents errors or biases from spreading across different memory levels. Meanwhile, it enables shared historical information to be independently represented at various levels of abstraction. Experiments demonstrate strong navigation performance on VLN-CE benchmarks and effective instruction following on a real robot.
\end{abstract}

%% file: sec/1_intro.tex
\section{Introduction}
\label{sec:intro}

Vision-and-language navigation (VLN) \citep{vln,vlnce} requires an embodied agent to interpret natural-language instructions and sequentially interact with complex visual environments to reach a target destination. Unlike static vision-language understanding, navigation decisions are inherently history-dependent \citep{vlnbert}: the agent must continuously relate its current observation to previously visited locations, encountered landmarks, and completed sub-goals in order to infer its progress and determine the next action \citep{hamt}. This makes memory a fundamental component of VLN systems. An effective navigation memory should not only preserve useful historical information over long trajectories, but also selectively retain observations that remain relevant to future decisions \citep{memoir}, while avoiding excessive accumulation of repetitive visual content \citep{janusvln}.

\begin{figure}
    \centering
    \includegraphics[width=\linewidth]{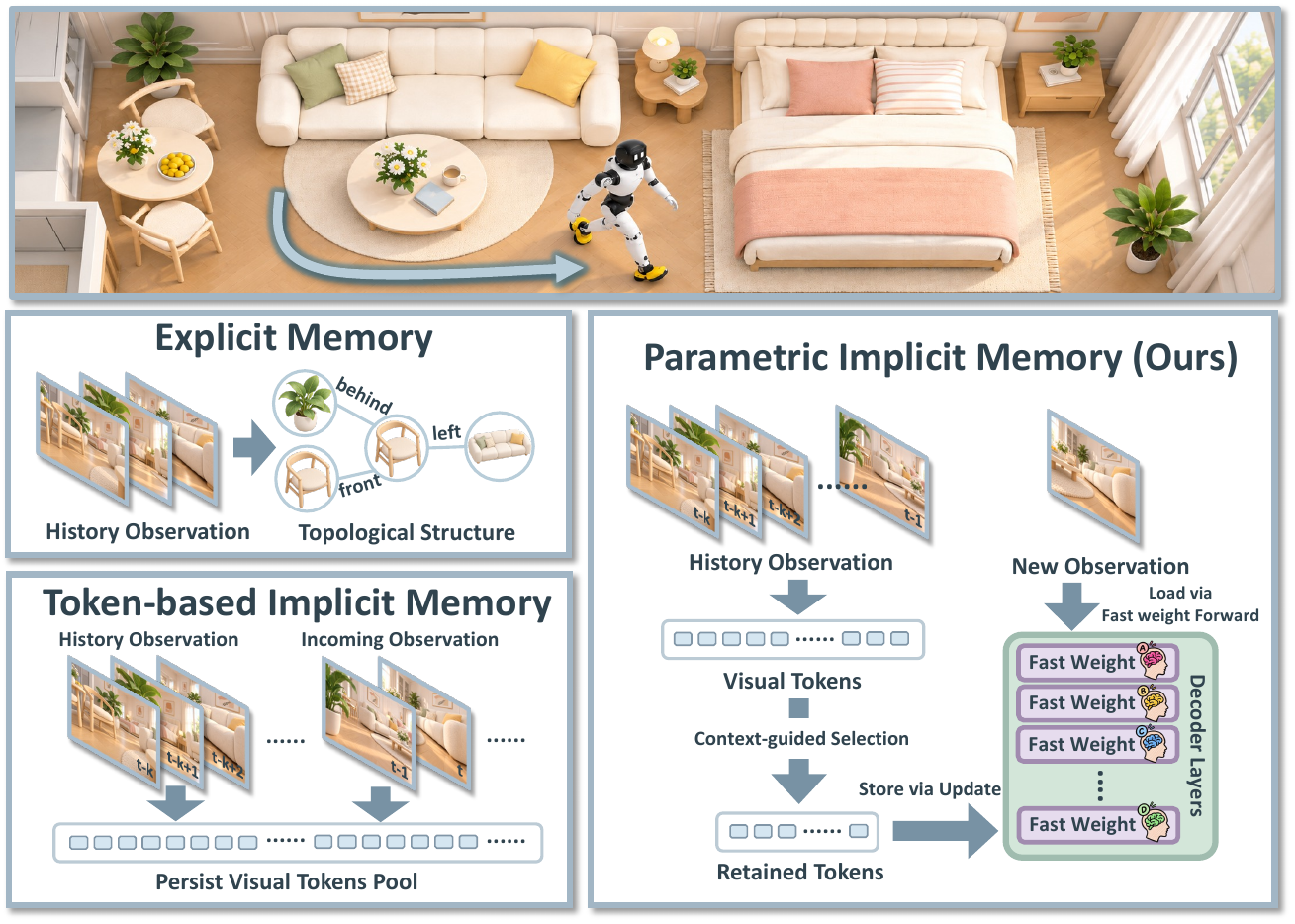}
    \caption{\textbf{From explicit to parametric memory in VLN.}
    \textbf{NavTTT} replaces explicit or token-based history with episode-specific parametric memory stored in fast weights.}
    \label{fig:teaser}
\end{figure}

Existing VLN methods maintain navigation history through either explicit \citep{hamt,duet,etpnav} or implicit memory \citep{vlnbert,krantz2023iterative,janusvln}. Explicit memory organizes past experience into structured representations, such as maps, graphs, or trajectories, making historical state directly accessible to the agent. Implicit memory instead compresses history into latent representations, typically as recurrent features or persistent tokens propagated across time. While less structured, these implicit memories remain fundamentally token-based: retaining more history still requires preserving, updating, or repeatedly attending to an expanding set of latent states. This dependence becomes increasingly problematic in long-horizon navigation, where successive observations are highly redundant and only a small fraction of past information may remain relevant to future decisions. As a result, token-based memory must continuously balance historical coverage against a finite representational and computational budget.

This limitation motivates a different form of implicit memory, where past experience is stored in the model state rather than kept as persistent representations, as illustrated in Fig.~\ref{fig:teaser}. Test-time training (TTT) \citep{sun2020ttt,sun2025ttt} provides a mechanism for doing so by updating fast weights with incoming observations, allowing past information to persist in the model parameters \citep{titans}. However, applying TTT directly to VLN is not straightforward. Navigation observations are highly redundant: nearby steps often produce similar visual inputs, while only a few observations remain useful for later decisions. Which observations matter also depends on the instruction, the agent's current location, and its progress along the route. Simply writing every observation into fast weights can therefore fill the memory with repeated or short-lived visual information and make important landmarks harder to preserve. In addition, navigation history needs to support different stages of decision making, while a generic fast-weight memory does not specify how historical information should be used across the navigation model. TTT-based memory for VLN therefore needs to address two key questions: what information should be stored, and how should that information be used during navigation.

To address these challenges, we introduce \textbf{Navigation-Conditioned Test-Time Training (NavTTT)}, a parametric memory framework that selectively internalizes navigation experience into episode-specific fast weights. NavTTT conditions memory formation on global navigation context, prioritizing visual evidence relevant to the instruction and navigation state while suppressing redundant content. It further integrates linguistic and spatial cues into the selected evidence, grounding memory in visual appearance and navigation context. To make accumulated experience accessible at multiple stages of navigation reasoning, NavTTT maintains independent fast-weight memories across decoder depths. Each memory transforms the selected evidence through its own association encoders and accumulates it via separate online updates. During subsequent decisions, hidden representations at each depth query the associated memory, integrating historical information with ongoing decoder computation. Together, adaptive memory formation and hierarchical memory integration allow navigation history to persist beyond limited visual context and inform subsequent decisions through layer-specific encoding and retrieval.

We evaluate NavTTT on the R2R-CE \citep{vln} and RxR-CE \citep{ku2020} benchmarks to examine the effectiveness of adaptive parametric memory for vision-and-language navigation. NavTTT achieves 66.7\% SR and 59.4\% SPL on R2R-CE, and 63.6\% SR and 52.7\% SPL on RxR-CE, consistently outperforming representative VLN baselines. Controlled ablations demonstrate that persistent fast-weight memory, online episode-level adaptation, and adaptive evidence selection each contribute substantially to navigation performance. Together, these results show that selectively internalizing navigation history through fast-weight updates provides an effective mechanism for preserving and exploiting long-horizon experience beyond the active visual context.

\begin{itemize}
    \item We introduce NavTTT, a parametric memory framework for vision-and-language navigation that internalizes observations into episode-specific fast weights through test-time training, making accumulated experience accessible beyond the explicit visual context.
    \item We propose navigation-conditioned memory formation that jointly considers task relevance and visual redundancy when selecting evidence, and incorporates linguistic and spatial cues to construct grounded representations for memory updates.
    \item We develop a hierarchical memory architecture that independently encodes and accumulates shared historical evidence across decoder depths. Layer-specific queries retrieve this information during subsequent decisions, integrating persistent memory into multiple stages of navigation reasoning.
\end{itemize}

%% file: sec/2_related_work.tex
\section{Related Work}
\label{sec:formatting}

\subsection{Vision-Language Navigation and Memory}

Vision-language navigation (VLN) has evolved from vision-language pre-training and sequential history modeling toward foundation-model-based navigation. Early approaches leverage large-scale pre-training \citep{prevalent} or explicitly model navigation history, such as recurrent state propagation in VLN-BERT \citep{vlnbert} and long-horizon trajectory encoding in HAMT \citep{hamt}. DUET \citep{duet} further constructs an online topological map to couple global planning with local grounding. More recently, NavGPT \citep{navgpt} and NaviLLM \citep{navillm} explore LLM-based navigation reasoning, while NaVid \citep{navid}, Uni-NaVid \citep{uninavid}, and NaVILA \citep{navila} extend VLMs toward end-to-end, unified, and real-world navigation. Our work complements this line by focusing specifically on how historical observations are retained and exploited during navigation.

Recent VLN systems have consequently developed increasingly sophisticated memory mechanisms. StreamVLN \citep{streamvln} maintains a slow-updating compressed visual context alongside a fast streaming context, while NavMorph \citep{navmorph} introduces contextual evolution memory within a self-evolving world model. JanusVLN \citep{janusvln} maintains separate spatial-geometric and visual-semantic histories as compact implicit memories, whereas HiMemVLN \citep{himemvln} organizes navigation history through a hierarchical memory system. Memoir \citep{memoir} further studies retrieval of accumulated environmental and behavioral experience across navigation episodes. Unlike these approaches, NavTTT selectively writes navigation-relevant observations into online-adapted fast weights, treating the model parameters themselves as persistent memory during an episode.

\subsection{Parametric Memory and Test-Time Training}
Test-time training (TTT) was originally introduced for self-supervised adaptation on unlabeled test samples \citep{sun2020ttt}, but has recently been repurposed as a sequence-modeling mechanism in which fast weights serve as recurrent state \citep{sun2025ttt}. This view is closely related to earlier fast-weight formulations of linear attention: \citet{schlag2021linear} interpret linear Transformers as fast-weight programmers that store key--value associations in rapidly changing weights. Modern TTT generalizes this perspective by allowing a learnable, potentially nonlinear model to serve as the evolving parameter state. Titans \citep{titans} develops such online neural memory for long-context modeling, while LaCT \citep{tttdoneright} enables large nonlinear fast-weight states through efficient large-chunk updates.

Recent work further explores fast weights as explicit parametric memory for adaptive inference and agents. TMEM \citep{tmem}, for example, consolidates interaction experience into lightweight online-updated parameters rather than relying solely on prompt-space memory. NavTTT adopts LaCT's nonlinear fast-weight formulation as its underlying memory operator, but focuses on a different question: how TTT-based memory should be organized for embodied navigation. Specifically, we study \emph{what to write} into fast weights from redundant visual streams and \emph{where to write} this information across the navigation model.

%% file: sec/3_method.tex
\section{Methodology}

\subsection{Preliminary}
\noindent \textbf{Navigation task.}
Vision-and-language navigation in continuous environments (VLN-CE) tasks an embodied agent with navigating to a target destination by following a natural-language instruction $x$ via sequential interactions with the environment. At each decision step $t$, the agent observes an egocentric RGB image $I_t$, maintains a visual history \(\mathcal{H}_t\) of up to 7 sampled frames, and receives an episode-relative pose $s_t=(p_t,\sin\psi_t,\cos\psi_t)$, where $p_t$ denotes its 2D position and $\psi_t$ its heading. These inputs are denoted by $\mathcal X_t=(x,\mathcal H_t,I_t,s_t,t)$. Conditioned on the instruction and all available observations, the navigation policy predicts an action sequence $A_t$ of up to 4 primitive actions drawn from the discrete action space \(\mathcal{A} = \{\textsc{Forward}, \textsc{Left}, \textsc{Right}, \textsc{Stop}\}\), where $t$ indexes decision cycles rather than individual actions. An episode terminates upon the \textsc{Stop} action or exhaustion of the predefined interaction budget. The core objective is to reach the target location efficiently while faithfully adhering to the given instruction.

\noindent \textbf{Test-Time Training.}
In the fast-weight formulation, test-time training (TTT)
enables online adaptation through a neural memory $f_W$,
whose parameters are updated as new observations arrive,
while the pretrained backbone remains fixed during inference.
Given a query $q_t$ and selected key--value pairs $(k_i,v_i)$,
NavTTT adopts a causal read-before-write order.
In the basic gradient-descent form:
\begin{equation}
\begin{aligned}
    o_t &= f_{W_{t-1}}(q_t),\\
    W_t &= W_{t-1}
    -\eta_t\nabla\mathcal L_{\mathrm{mem}}(W_{t-1}),
\end{aligned}
\label{eq:ttt_preliminary}
\end{equation}
where $W_t$ denotes the fast weights, $\eta_t$ is the update
rate, and the gradient is taken with respect to $W$.
Following LaCT~\citep{tttdoneright}, the memory objective
is the negative dot-product loss:
\begin{equation}
    \mathcal L_{\mathrm{mem}}(W)
    =-\sum_i m_i\left\langle f_W(k_i),v_i\right\rangle,
    \label{eq:navttt_memory_loss}
\end{equation}
where $m_i$ is the binary selection mask and
$\langle\cdot,\cdot\rangle$ denotes the inner product.
This objective strengthens key--value associations by
increasing the agreement between the memory response
$f_W(k_i)$ and its associated value $v_i$.
Layer and head indices are omitted; $M_t$ denotes the
complete memory state across layers.

\begin{figure*}[t]
    \centering
    \includegraphics[width=0.95\textwidth]{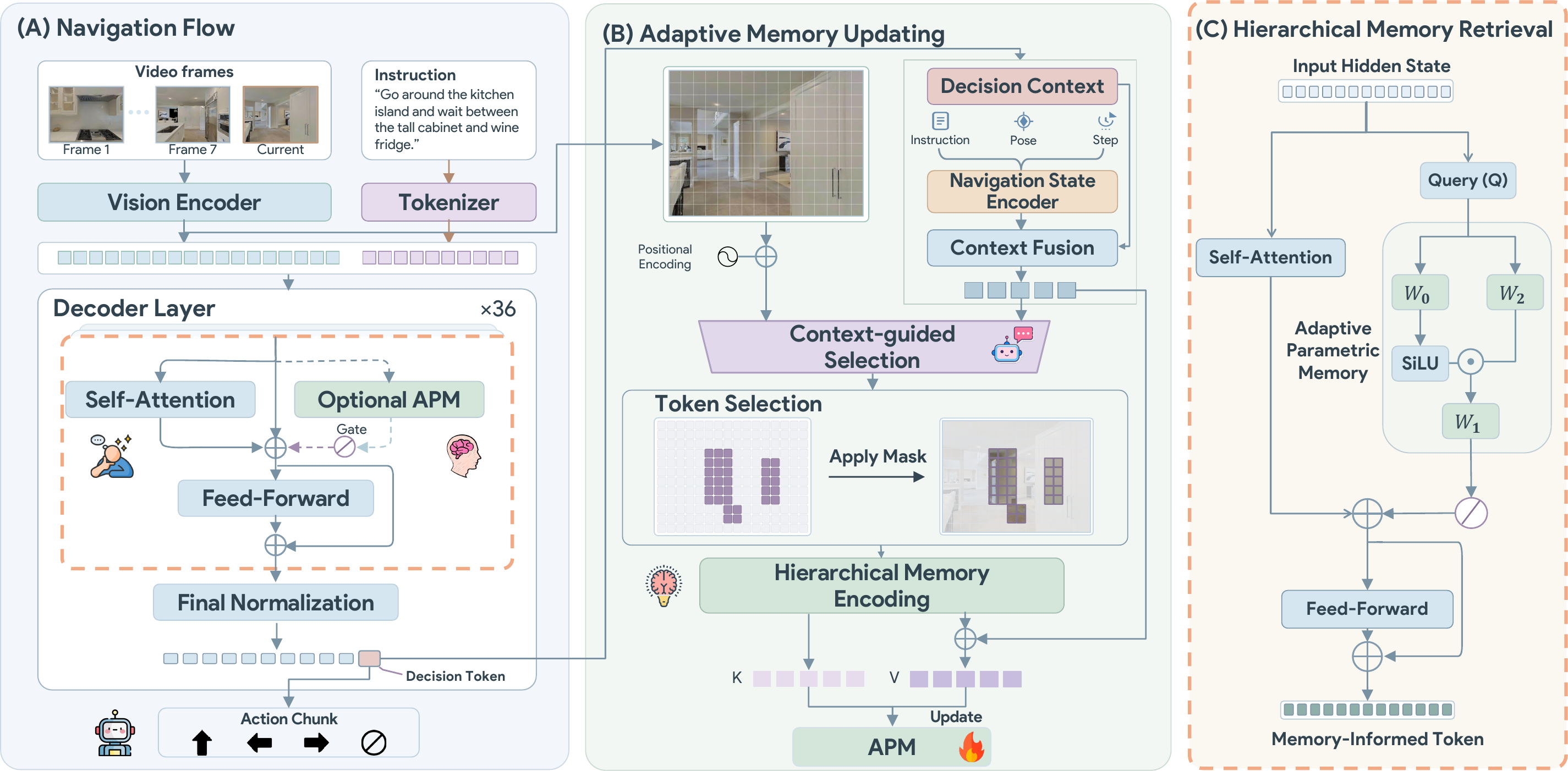}
    \caption{\textbf{Overview of NavTTT.}
    (A) The agent predicts low-level action chunks from language instructions and egocentric visual observations.
    (B) Navigation-relevant visual tokens are selectively encoded to update hierarchical fast weights online.
    (C) At subsequent steps, decoder representations query the adapted fast weights across layers, allowing accumulated experience to inform current perception and action prediction.}
    \label{fig:placeholder}
\end{figure*}

\subsection{Informative Token Selection}
\label{sec:token_selection}

\noindent \textbf{Task relevance and visual diversity.}
The relevance of a visual region depends on the navigation instruction and the agent's current state. However, selecting tokens only by relevance may include repeated visual content from the same frame, leaving fewer tokens for other useful regions under a fixed write budget. NavTTT therefore combines context-guided relevance scoring with intra-frame redundancy suppression to select relevant and diverse visual tokens for APM updates. The selection mask controls memory writing, while all visual tokens remain available to the navigation backbone.

\paragraph{Context-Guided Token Selection.} 
Given a visual observation \(I\), raw visual patch features \(e_i = \operatorname{sg}\left([E_{\text{vis}}(I)]_i\right)\) are extracted via a frozen vision encoder \(E_{\text{vis}}(\cdot)\), and augmented with Fourier positional encodings to generate position-aware visual tokens \(u_i\). In parallel, the Navigation State Encoder (NSE) integrates and encodes non-visual navigation signals, including the navigation instruction \(g\), agent pose \(s\), and decision step \(k\), into a compact explicit state variable \(c\):
\begin{equation}
    c = \text{NSE}(g, s, k).
\end{equation}
As illustrated in Fig.~\ref{fig:placeholder}(B), \(c\) is fused with the decoder's high-level decision token \(d\) to construct a global guidance context \(z\), expressed as:
\begin{equation}
z = W_d \operatorname{RMSNorm}\left(\operatorname{sg}(d)\right) + W_c c,
\end{equation}
where \(W_d\) and \(W_c\) denote learnable projection matrices.

This global guidance context serves as a conditional vector that fuses with the features of each visual token along the channel dimension, thereby modulating the feature representation. Based on the accumulated navigation state, it regulates the assessment of visual input relevance, establishing a closed-loop mechanism wherein visual observations are filtered adaptively according to the task state. The relevance score \(r_i\) for the i-th token is computed as:
\begin{equation}
r_i = w_s^\top \operatorname{SiLU}\left(\operatorname{LayerNorm}\left(W_u u_i + z\right)\right),
\label{eq:navttt_relevance_scoring}
\end{equation}
where \(W_u\) is a visual projection matrix and \(w_s\) is a learnable scoring vector.

Relying solely on independent scoring fails to suppress visual redundancy, as multiple similar patches may occupy limited memory capacity. To address this, we introduce an incremental diversity selection step. Starting from an empty set, we select tokens one by one, penalizing candidate tokens that are visually similar to those already chosen. 	We provide the full token selection procedure in the supplementary material.

\subsection{Fast-Weight Memory for Navigation}
\label{sec:navttt_decoding}

NavTTT extends the limited visual context capacity by persisting selected task-relevant visual evidence via Adaptive Parametric Memory (APM). Following the LaCT~\citep{tttdoneright} framework, each memory-augmented decoder layer maintains an independent nonlinear fast-weight mapping to encode accumulated key-value associations. During navigation, this persistent memory complements the self-attention pathway over instructions and current observations, connecting current perception with historical experience from earlier decisions.

\subsubsection{Contextual Association Encoding}

Given the normalized visual features of each selected region, a layer-specific adapter produces $a_i$. The key and value representations at decoder layer $\ell$ are defined as:
\begin{equation}
\begin{aligned}
k_i &= \mathcal{T}_K\left(W_K^\ell a_i\right), \\
v_i &= W_V^\ell a_i
+ W_L^\ell \xi_i
+ W_z^\ell z,
\end{aligned}
\label{eq:navttt_key_value}
\end{equation}
with the superscript $\ell$ indicating layer-specific projections, and $\mathcal{T}_K$ applying normalization alongside positional encoding. Specifically, the key encodes visual-positional cues for retrieval, whereas the value aggregates visual appearance $a_i$, spatial context $\xi_i$ (patch coordinates and relative agent poses), and the fused navigation state $z$. Ultimately, this formulation grounds the selected visual evidence in its corresponding spatial and task-specific context.

\subsubsection{Hierarchical Retrieval and Memory Fusion}
\label{sec:navttt_retrieval}

NavTTT combines visual context with persistent
experience through a hierarchical dual-path decoder.
Self-attention is retained throughout the 36-layer backbone,
while parallel APM branches are placed at layers
$\{5,11,17,23,29,35\}$, using zero-based indices.
These branches provide memory access at
different decoder depths.
Each augmented layer maintains independent fast weights,
allowing historical evidence to be encoded and retrieved
through layer-specific representations.

Each APM branch contains four independent memory heads.
Within each head, a query $q$ derived from the decoder
hidden state retrieves stored experience through a
nonlinear SwiGLU mapping:
\begin{equation}
    f_{W^\ell}(q)
    =
    W_1^\ell
    \left[
        \operatorname{SiLU}(W_0^\ell q)
        \odot
        (W_2^\ell q)
    \right],
    \label{eq:navttt_fast_network}
\end{equation}
where $W^\ell=\{W_0^\ell,W_1^\ell,W_2^\ell\}$ denotes
the fast weights, and $\odot$ denotes elementwise
multiplication.
Head indices are omitted for simplicity.
This mapping provides access to historical evidence
after its source frames leave the active visual context. At decision step $t$, the self-attention and memory
pathways are combined through a residual connection:
\begin{equation}
    \widetilde{h}_j
    =
    h_j
    +
    \operatorname{SelfAttn}(H)_j
    +
    \gamma_j
    \operatorname{APM}(h_j;W_{t-1}^\ell),
    \label{eq:navttt_decoder_residual}
\end{equation}
where $h_j$ is the hidden state at position $j$ in
the current layer's input sequence $H$.
The APM operator combines the memory-head outputs and
projects them into the decoder hidden space.
The gate $\gamma_j$ is fixed to $1$ at decision and
action positions.
For preselected visual tokens in layers $\{11,17,23\}$,
it is computed from the corresponding normalized
pre-decoder visual feature through a learned projection
followed by a sigmoid.
The gate is set to $0$ elsewhere.
These visual read positions are determined during prompt
processing, before the final decision representation
is available.
Normalization, headwise scaling, and output projection
are included in the corresponding operators.
The fused state $\widetilde{h}_j$ is then passed to the
standard feed-forward block.

The state $W_{t-1}^\ell$ contains experience accumulated
before decision $t$ and remains unchanged during current
prompt processing and action generation.
Once the final decision representation is available,
the selected key-value pairs from the current observation
produce a pending fast-weight update using gradients
of the memory association loss $\mathcal L_{\mathrm{mem}}$.
After the current action chunk is finalized,
the updated state is committed as $W_t^\ell$ and becomes
available for retrieval at decision $t+1$.

%% file: sec/4_experiment.tex
\section{Experiments}
\label{sec:experiments}

\subsection{Experimental Setup}
\label{sec:exp_setup}

\noindent \textbf{Simulation environments and metrics.}
Following standard evaluation protocols for continuous vision-and-language navigation~\citep{streamvln,janusvln}, we evaluate NavTTT on the R2R-CE and RxR-CE benchmarks~\citep{vlnce,vln,ku2020}. Both benchmarks require agents to follow natural-language instructions and navigate continuous Matterport3D~\citep{Matterport3D} environments via low-level actions. Experiments are evaluated on the R2R-CE Val-Unseen split and the RxR-CE Val-Unseen split. R2R-CE performance is reported using NE, OS, SR, and SPL,
while RxR-CE performance is reported using NE, SR, SPL,
and nDTW. SR and SPL serve as primary metrics for task completion and navigation efficiency.

\begin{figure*}[t]
    \centering
    \includegraphics[width=0.85\textwidth]{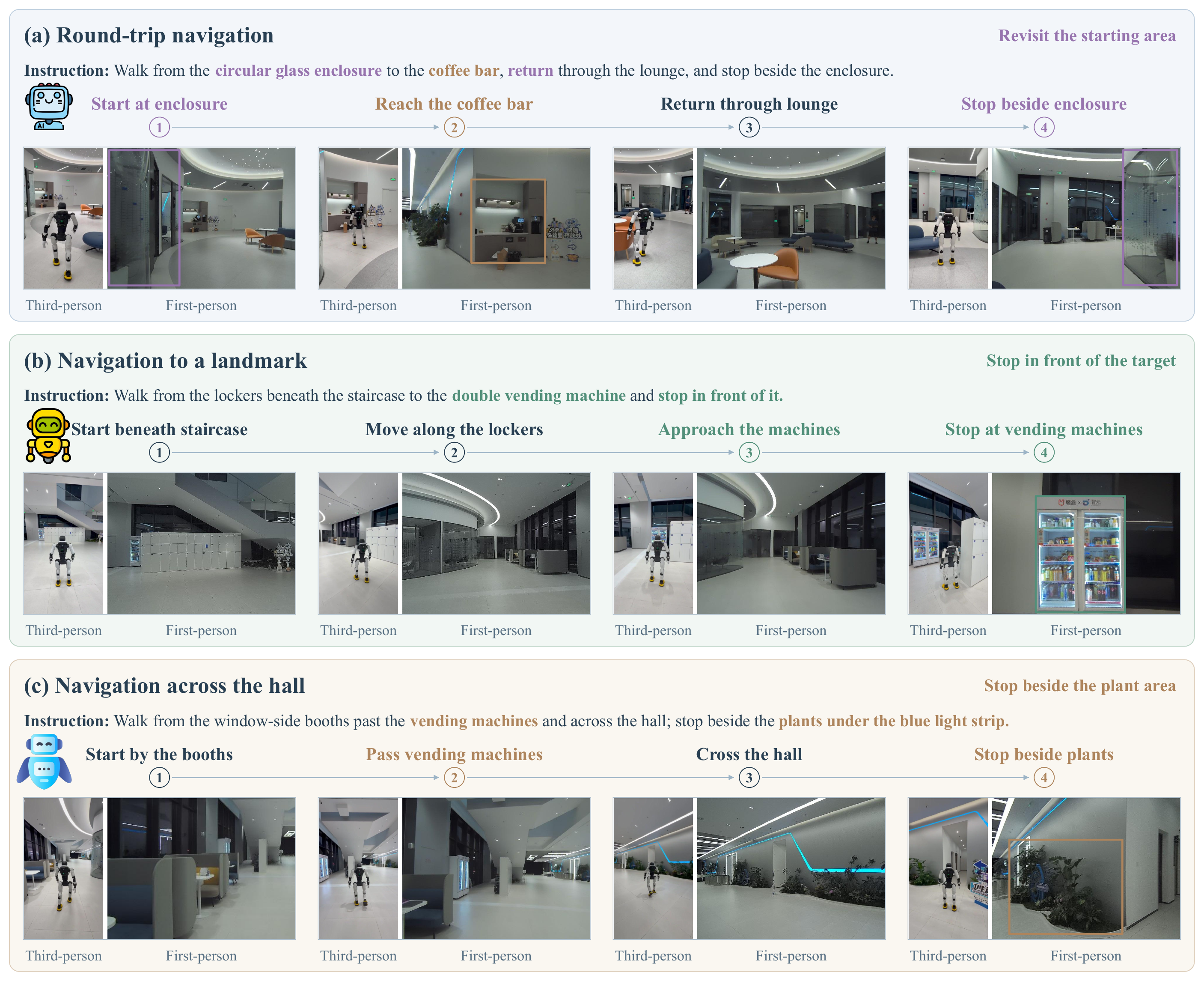}
    \caption{\textbf{Real-world qualitative results of NavTTT.}
    Representative instruction-following episodes on the real
    robot illustrate NavTTT's navigation behavior in physical
    environments, guided by natural-language instructions and
    egocentric RGB observations.}
    \label{fig:realbot}
\end{figure*}

\noindent \textbf{Real-world evaluation setup.}
We deploy NavTTT on a humanoid robot (shown in Figure~\ref{fig:realbot}) to assess the real-world deployment capability and Sim-to-Real generalization of its learned memory mechanism in continuous navigation tasks. The robot perceives the environment through a forward-facing RGB camera, receiving natural-language instructions alongside streaming egocentric visual observations. NavTTT runs on an off-board inference server equipped with an NVIDIA RTX 4090 GPU, transmitting predicted low-level navigation actions to the robot for execution. No real-world trajectories are used for fine-tuning, enabling a direct assessment of sim-to-real generalization.

\noindent \textbf{Implementation details.}
NavTTT is initialized from
Qwen3-VL-8B~\citep{bai2025qwen3vltechnicalreport}, with APM
integrated from the beginning of joint training.
The language backbone, multimodal projector, and APM modules,
including the learned fast-weight initialization, are jointly
fine-tuned, while the vision encoder remains frozen.
Joint training runs for one epoch using AdamW with a weight decay of
$10^{-2}$ and a cosine learning-rate schedule with a warmup
ratio of $0.03$, consuming approximately 1,910 GPU hours on
NVIDIA H20 GPUs.
The peak learning rates are $1\times10^{-5}$ for the language
backbone and $5\times10^{-6}$ for the multimodal projector.
The architectural ablations in Section~\ref{sec:ablation}
follow the same joint-training protocol as the main model.
We detail the training objective and inference-time fast-weight updates in the supplementary material.

\providecommand{\NavTTTCheck}{$\checkmark$}
\providecommand{\NavTTTVenue}[1]{{\scriptsize[#1]}}

\begin{table*}[!t]
    \centering
    \caption{
    Comparison among different methods on R2R-CE and
    RxR-CE Val-Unseen.
    Pano., Odo., Depth, and S.RGB denote panoramic input,
    odometry/pose, depth, and a single RGB stream, respectively.
    Baseline external-data counts follow \citep{janusvln},
    with $\mathrm{K}=10^3$ image--action or QA pairs;
    generic foundation-model pretraining is excluded.
    ``--'' denotes an unavailable count or a metric without
    a verified result for the corresponding split.
    $0\mathrm{K}$ denotes a reported zero count.
    NE is measured in meters; other metrics are percentages.
    Best values are bold.
    }
    \label{tab:vlnce_main}

    \small
    \setlength{\tabcolsep}{1.8pt}
    \renewcommand{\arraystretch}{1.05}

    \resizebox{\textwidth}{!}{%
    \begin{tabular}{@{}l*{4}{c}*{8}{r}r@{}}
        \toprule
        \multirow{2}{*}{Method}
        & \multicolumn{4}{c}{Observation}
        & \multicolumn{4}{c}{R2R-CE Val-Unseen}
        & \multicolumn{4}{c}{RxR-CE Val-Unseen}
        & \multicolumn{1}{c}{Training} \\

        \cmidrule(lr){2-5}
        \cmidrule(lr){6-9}
        \cmidrule(lr){10-13}
        \cmidrule(lr){14-14}

        & Pano. & Odo. & Depth & S.RGB
        & NE$\downarrow$ & OS$\uparrow$
        & SR$\uparrow$ & SPL$\uparrow$
        & NE$\downarrow$ & SR$\uparrow$
        & SPL$\uparrow$ & nDTW$\uparrow$
        & \multicolumn{1}{c}{External Data} \\

        \midrule

        % Panoramic-input methods.
        HPN+DN \NavTTTVenue{ICCV21}~\citep{hpn}
        & \NavTTTCheck & \NavTTTCheck & \NavTTTCheck &
        & 6.31 & 40.0 & 36.0 & 34.0
        & -- & -- & -- & --
        & -- \\

        CMA \NavTTTVenue{CVPR22}~\citep{hong2022bridging}
        & \NavTTTCheck & \NavTTTCheck & \NavTTTCheck &
        & 6.20 & 52.0 & 41.0 & 36.0
        & 8.76 & 26.5 & 22.1 & 47.0
        & -- \\

        Sim2Sim \NavTTTVenue{ECCV22}~\citep{sim2sim}
        & \NavTTTCheck & \NavTTTCheck & \NavTTTCheck &
        & 6.07 & 52.0 & 43.0 & 36.0
        & -- & -- & -- & --
        & -- \\

        VLN$\circlearrowright$BERT
        \NavTTTVenue{CVPR22}~\citep{hong2022bridging}
        & \NavTTTCheck & \NavTTTCheck & \NavTTTCheck &
        & 5.74 & 53.0 & 44.0 & 39.0
        & 8.98 & 27.0 & 22.6 & 46.7
        & -- \\

        Ego$^{2}$-Map \NavTTTVenue{ICCV23}~\citep{ego2map}
        & \NavTTTCheck & \NavTTTCheck & \NavTTTCheck &
        & 5.54 & 56.0 & 47.0 & 41.0
        & -- & -- & -- & --
        & -- \\

        DreamWalker \NavTTTVenue{ICCV23}~\citep{dreamwalker}
        & \NavTTTCheck & \NavTTTCheck & \NavTTTCheck &
        & 5.53 & 59.0 & 49.0 & 44.0
        & -- & -- & -- & --
        & -- \\

        GridMM \NavTTTVenue{ICCV23}~\citep{wang2023gridmmgridmemorymap}
        & \NavTTTCheck & \NavTTTCheck & \NavTTTCheck &
        & 5.11 & 61.0 & 49.0 & 41.0
        & -- & -- & -- & --
        & -- \\

        InstructNav \NavTTTVenue{CoRL24}~\citep{instructnav}
        & \NavTTTCheck & \NavTTTCheck & \NavTTTCheck &
        & 6.89 & -- & 31.0 & 24.0
        & -- & -- & -- & --
        & -- \\

        COSMO \NavTTTVenue{ICCV25}~\citep{cosmo}
        & \NavTTTCheck & & &
        & -- & 56.0 & 47.0 & 40.0
        & -- & -- & -- & --
        & -- \\

        AO-Planner \NavTTTVenue{AAAI25}~\citep{aoplanner}
        & \NavTTTCheck & & \NavTTTCheck &
        & 5.55 & 59.0 & 47.0 & 33.0
        & 7.06 & 43.3 & 30.5 & 50.1
        & -- \\

        g3D-LF (Pano.) \NavTTTVenue{CVPR25}~\citep{wang2024g3dlf}
        & \NavTTTCheck & \NavTTTCheck & \NavTTTCheck &
        & 4.53 & 68.0 & 61.0 & 52.0
        & -- & -- & -- & --
        & -- \\

        \midrule

        MapNav \NavTTTVenue{ACL25}~\citep{mapnav}
        & & \NavTTTCheck & \NavTTTCheck & \NavTTTCheck
        & 4.93 & 53.0 & 39.7 & 37.2
        & 7.62 & 32.6 & 27.7 & 43.5
        & -- \\

        g3D-LF (Mono.) \NavTTTVenue{CVPR25}~\citep{wang2024g3dlf}
        & & \NavTTTCheck & \NavTTTCheck & \NavTTTCheck
        & 5.70 & 59.5 & 47.2 & 34.6
        & -- & -- & -- & --
        & -- \\

        % RxR results are omitted because the split attribution
        % differs across published comparison tables.
        Seq2Seq \NavTTTVenue{ECCV20}~\citep{vlnce}
        & & & \NavTTTCheck & \NavTTTCheck
        & 7.77 & 37.0 & 25.0 & 22.0
        & -- & -- & -- & --
        & -- \\

        NaVid-4D \NavTTTVenue{ICRA25}~\citep{navid4d}
        & & & \NavTTTCheck & \NavTTTCheck
        & 5.99 & 55.7 & 43.8 & 37.1
        & -- & -- & -- & --
        & -- \\

        NavMorph \NavTTTVenue{ICCV25}~\citep{navmorph}
        & & & \NavTTTCheck & \NavTTTCheck
        & 5.75 & 56.9 & 47.9 & 33.2
        & 8.85 & 30.8 & 22.8 & 44.2
        & -- \\

        \midrule
        Sim2Real \NavTTTVenue{CoRL24}~\citep{simtoreal}
        &  &  &  & \NavTTTCheck
        & 5.95 & 55.8 & 44.9 & 30.4
        & 8.79 & 25.5 & 18.1 & --
        & $0\mathrm{K}$ \\

        % Single-RGB methods and external-data configurations.
        NaVid \NavTTTVenue{RSS24}~\citep{navid}
        & & & & \NavTTTCheck
        & 5.47 & 49.1 & 37.4 & 35.9
        & 8.41 & 23.8 & 21.2 & --
        & $953\mathrm{K}$ \\

        Uni-NaVid \NavTTTVenue{RSS25}~\citep{uninavid}
        & & & & \NavTTTCheck
        & 5.58 & 53.3 & 47.0 & 42.7
        & 6.24 & 48.7 & 40.9 & --
        & $3577\mathrm{K}$ \\

        NaVILA \NavTTTVenue{RSS25}~\citep{navila}
        & & & & \NavTTTCheck
        & 5.22 & 62.5 & 54.0 & 49.0
        & 6.77 & 49.3 & 44.0 & 58.8
        & $13132\mathrm{K}$ \\

        % Use the version reported in JanusVLN for consistency
        % between benchmark results and external-data counts.
        StreamVLN \NavTTTVenue{ICRA26}~\citep{streamvln}
        & & & & \NavTTTCheck
        & 4.98 & 64.2 & 56.9 & 51.9
        & 6.22 & 52.9 & 46.0 & 61.9
        & $\sim 26330\mathrm{K}$ \\

        JanusVLN \NavTTTVenue{ICLR26}~\citep{janusvln}
        & & & & \NavTTTCheck
        & 4.78 & 65.2 & 60.5 & 56.8
        & 6.06 & 56.2 & 47.5 & 62.1
        & $10692\mathrm{K}$ \\

        \midrule

        \textbf{NavTTT}
        & & \NavTTTCheck & & \NavTTTCheck
        & \textbf{3.88} & \textbf{73.1}
        & \textbf{66.7} & \textbf{59.4}
        & \textbf{4.25} & \textbf{63.6}
        & \textbf{52.7} & \textbf{71.0}
        & $2974\mathrm{K}^{\ddagger}$ \\

        \bottomrule
    \end{tabular}%
    }

    % \vspace{1pt}
    \begin{minipage}{\textwidth}
        \footnotesize

        \noindent
$\ddagger$ NavTTT uses 2,974,963 additional DAgger image--action pairs collected from 30,497 trajectories: 780,222
  pairs from 10,819 R2R trajectories and 2,194,741 pairs from 19,678 RxR trajectories.
    \end{minipage}
\end{table*}

\subsection{Main Results}
\label{sec:main_results}

\noindent \textbf{Results on VLN-CE benchmarks.}
Table~\ref{tab:vlnce_main} compares NavTTT with representative VLN-CE methods across observation modalities and training setups. On R2R-CE Val-Unseen, NavTTT achieves a leading \textbf{66.7\%} SR and \textbf{59.4\%} SPL, outperforming JanusVLN by \textbf{6.2} SR / \textbf{2.6} SPL and StreamVLN by \textbf{9.8} SR / \textbf{7.5} SPL. These improvements demonstrate how our selective memory pipeline resolves perceptual ambiguities. Context-guided selection isolates task-relevant visual regions, whereas intra-frame redundancy suppression ensures maximal feature diversity within each fast-weight update. NavTTT parametrically registers these distilled historical anchors into fast weights and queries them alongside streaming observations, enabling robust long-horizon contextual reasoning. On RxR-CE Val-Unseen, which presents long-horizon trajectories and detailed instructions, NavTTT reaches \textbf{63.6\%} SR, \textbf{52.7\%} SPL, and \textbf{71.0\%} nDTW. Here, APM retrieves long-range evidence that has exited the active visual window, while the self-attention pathway preserves local grounding, ensuring smooth continuity across extended navigation steps. 

Figure~\ref{fig:realbot} presents representative
instruction-following episodes on the real robot.
Across 40 real-world navigation trials, NavTTT successfully
completes 11, achieving a success rate of \textbf{27.5\%}.
During execution, the agent processes successive egocentric
observations and predicts short action chunks for the robot
to execute.
Its fast-weight memory is updated online with newly observed
evidence, making accumulated context available to subsequent
decisions.
This evaluation complements the simulation benchmarks and
provides a quantitative assessment of NavTTT on a physical
platform, while highlighting substantial room for improving
real-world navigation reliability.

The larger SPL gain over JanusVLN on RxR-CE (5.2 percentage points) than on R2R-CE (2.6 percentage points) is consistent with the motivation to retain historical evidence over longer trajectories. The accompanying improvement in nDTW indicates better alignment with reference routes, complementing the gains in task completion. A possible explanation is that persistent memory makes previously observed landmarks and their navigation context available when recent frames alone provide insufficient information. Context-guided selection and redundancy suppression determine which evidence enters this memory, while hierarchical retrieval allows the retained information to support subsequent decisions. NavTTT obtains these results using approximately 2.97 million additional image--action training pairs. This reported data requirement should be interpreted alongside differences in observation modalities and training settings across methods. We provide additional qualitative results and failure cases in the supplementary material.
\subsection{Ablation Studies}
\label{sec:ablation}

We examine the contributions of fast-weight memory,
online adaptation, and navigation-specific memory design
on R2R-CE and RxR-CE Val-Unseen. We use
Qwen3-VL-8B~\citep{bai2025qwen3vltechnicalreport} and
Qwen2.5-VL-7B-Instruct~\citep{qwen2.5}. Within each
backbone, the architectural variants follow the main
joint-training protocol with matched backbone initialization,
observation inputs, training data, and optimization budget.
The static-memory control uses the full model checkpoint
and holds its learned initial fast state fixed throughout
evaluation.

\begin{table*}[t]
    \centering
    \caption{Memory ablations on R2R-CE and RxR-CE Val-Unseen.
    NE is measured in meters. OS, SR, SPL, and nDTW are
    reported on a percentage scale. Bold marks the best
    result within each backbone and benchmark.}
    \label{tab:memory_formation}
    \small
    \setlength{\tabcolsep}{5pt}
    \begin{tabular}{@{}cl*{8}{c}@{}}
        \toprule
        & & \multicolumn{4}{c}{R2R-CE Val-Unseen}
          & \multicolumn{4}{c}{RxR-CE Val-Unseen} \\
        \cmidrule(lr){3-6}
        \cmidrule(lr){7-10}
        ID & Variant
           & NE$\downarrow$ & OS$\uparrow$
           & SR$\uparrow$ & SPL$\uparrow$
           & NE$\downarrow$ & SR$\uparrow$
           & SPL$\uparrow$ & nDTW$\uparrow$ \\
        \midrule
        \multicolumn{10}{@{}l}{\textit{Qwen3-VL-8B}} \\
        \addlinespace[2pt]
        A0 & NavTTT (full)
           & \textbf{3.88} & \textbf{73.1}
           & \textbf{66.7} & \textbf{59.4}
           & \textbf{4.25} & \textbf{63.6}
           & 52.7 & \textbf{71.0} \\
        A1 & No fast-weight memory
           & 4.55 & 68.7 & 62.4 & 56.5
           & 6.23 & 60.2 & 52.9 & 62.3 \\
        A2 & Static memory
           & 4.46 & 69.7 & 63.3 & 56.8
           & 5.62 & 61.3 & 52.5 & 66.9 \\
        A3 & Direct LaCT adaptation
           & 4.39 & 70.2 & 63.3 & 56.4
           & 5.24 & 62.3 & \textbf{53.7} & 68.3 \\
        \midrule
        \multicolumn{10}{@{}l}{\textit{Qwen2.5-VL-7B-Instruct}} \\
        \addlinespace[2pt]
        B0 & NavTTT (full)
           & \textbf{4.83} & \textbf{67.3}
           & \textbf{57.3} & \textbf{51.0}
           & \textbf{6.71} & \textbf{53.4}
           & \textbf{44.8} & \textbf{63.1} \\
        B1 & No fast-weight memory
           & 5.53 & 62.4 & 53.6 & 48.6
           & 7.66 & 50.7 & 43.0 & 61.4 \\
        B2 & Static memory
           & 4.91 & 63.6 & 56.4 & 50.3
           & 7.20 & 52.1 & 43.3 & 62.5 \\
        B3 & Direct LaCT adaptation
           & 4.97 & 64.2 & 55.7 & 49.7
           & 7.46 & 51.6 & 42.1 & 60.4 \\
        \bottomrule
    \end{tabular}
\end{table*}

\paragraph{Contribution of fast-weight memory.}
In Table~\ref{tab:memory_formation}, full NavTTT improves
SR over the memory-free controls (A1 and B1) by 4.3 and
3.4 percentage points on R2R-CE and RxR-CE with
Qwen3-VL-8B, and by 3.7 and 2.7 points with
Qwen2.5-VL-7B-Instruct. NE also decreases in all four
settings. The higher oracle success rates on R2R-CE
indicate that more trajectories reach the goal neighborhood
during navigation. On RxR-CE, nDTW improves with both
backbones, and the Qwen3-VL-8B model's SPL is 0.2 points
below its memory-free baseline.

\paragraph{Contribution of online adaptation.}
Across the four settings in Table~\ref{tab:memory_formation},
static memory improves SR by 0.9--2.8 percentage points over
the memory-free controls. To isolate the effect of online
writes, we compare A0 with A2 and B0 with B2 using the same
trained checkpoint within each pair. Enabling online writes
improves SR and SPL and reduces NE in all four settings.
On R2R-CE with Qwen3-VL-8B, online updates yield 61 additional
successes across 1,839 episodes, corresponding to a 3.32-point
increase in SR, and improve SPL by 2.6 points. On RxR-CE with
the same backbone, SR, SPL, and nDTW improve by 2.3, 0.2,
and 4.1 points, respectively. With Qwen2.5-VL-7B-Instruct,
the SR and SPL gains are 0.9 and 0.7 points on R2R-CE and
1.3 and 1.5 points on RxR-CE. On RxR-CE, the nDTW gains
with both backbones indicate closer alignment with the
reference trajectories when the agent incorporates new
observations into its fast state.

\paragraph{Navigation-specific memory design.}
Direct LaCT improves SR over the memory-free controls
in all four settings. Full NavTTT improves SR further
and reduces NE. On R2R-CE, its SPL gains over direct LaCT
are 3.0 points with Qwen3-VL-8B and 1.3 points with
Qwen2.5-VL-7B-Instruct. On RxR-CE, nDTW improves by
2.7 points with each backbone. SPL is 1.0 point lower
with Qwen3-VL-8B and 2.7 points higher with
Qwen2.5-VL-7B-Instruct. 

The LaCT comparison evaluates selective writing, spatial
and navigation context encoding, and multi-depth retrieval
as a combined design. At the same write budget, combining
relevance and within-frame diversity improves SR and SPL
over either criterion alone. At the same fast-state storage,
the six-layer shared configuration improves SR and SPL over
last-layer access. The supplementary material provides
details of both ablations, additional memory analyses,
and inference latency measurements.

%% file: sec/5_conclusion.tex
\section{Conclusion}

This paper presented NavTTT, a framework that uses fast-weight memory to retain historical information for vision-and-language navigation. The method combines navigation-conditioned token selection, contextual association encoding, and memory retrieval at multiple decoder layers. Token selection balances task relevance and visual diversity under a fixed write budget, while contextual encoding associates the selected visual features with spatial information and navigation state. The decoder combines self-attention over the available context with retrieval from episode-local memory, allowing earlier observations to inform subsequent decisions after their source frames leave the active visual window. During inference, the fast weights are updated through a self-supervised objective, while the backbone and other slow-weight parameters remain fixed. Experiments on R2R-CE and RxR-CE show improvements in navigation success and efficiency. We also evaluated the method on a real robot to examine its behavior outside simulation. These results suggest that selectively storing and retrieving historical evidence can complement the limited visual context available to the navigation model. Future work will investigate memory retention over longer trajectories and performance in environments with repeated landmarks or greater visual ambiguity.

%% file: supplmental_matrerial.tex
% CVPR 2027 supplementary material, compiled as a separate document.
% Use the same title as the main submission.
% Required project files: cvpr.sty and the four complete figure PDFs.
% Images may be placed beside this file or in figures/.
\graphicspath{{./}{figures/}}

\definecolor{cvprblue}{rgb}{0.21,0.49,0.74}
\def\confName{CVPR}
\def\confYear{2027}
\setlength{\textfloatsep}{12pt plus 2pt minus 2pt}
\setlength{\dbltextfloatsep}{12pt plus 2pt minus 2pt}
\setlength{\dblfloatsep}{10pt plus 2pt minus 2pt}
\widowpenalty=10000
\clubpenalty=10000
\hypersetup{pdftitle={Test-Time Evolution of Parametric Memory for Vision-and-Language Navigation -- Supplementary Material},pdfauthor={}}
\makeatletter
\setlength{\@fptop}{0pt}
\setlength{\@fpsep}{12pt}
\setlength{\@fpbot}{0pt plus 1fil}
\setlength{\@dblfptop}{0pt}
\setlength{\@dblfpsep}{12pt}
\setlength{\@dblfpbot}{0pt plus 1fil}
\makeatother

\maketitlesupplementary
\raggedbottom
\appendix
\setcounter{page}{1}
\setcounter{equation}{0}
\setcounter{figure}{0}
\setcounter{table}{0}
\renewcommand{\theequation}{S\arabic{equation}}
\renewcommand{\thefigure}{S\arabic{figure}}
\renewcommand{\thetable}{S\arabic{table}}

\section{Additional Method Details}
\label{app:method_details}

NavTTT uses navigation context to select visual evidence
for episode-local fast-weight memory. This section details
how selection accounts for redundancy. Section~\ref{app:training}
describes joint training and the inference-time update sequence.

\subsection{Intra-frame redundancy suppression}
\label{app:redundancy_suppression}

The relevance score $r_i$ evaluates each visual token
independently. Under a fixed write budget $K$, selecting
tokens solely by relevance can retain multiple patches
with similar visual content. We therefore construct a
selection set $\mathcal S$ incrementally. Starting from
$\mathcal S=\varnothing$, we repeatedly select a token using
\begin{equation}
\begin{aligned}
 i^* &= \arg\max_{i\notin\mathcal S}
 \left[r_i-\lambda_{\mathrm{div}}
 \max_{j\in\mathcal S}\cos(e_i,e_j)\right], \\
 \mathcal S &\leftarrow \mathcal S\cup\{i^*\}.
\end{aligned}
\label{eq:app_mmr_selection}
\end{equation}
Selection continues until $K$ tokens have been retained
or the candidates are exhausted. For an empty selection
set, the redundancy penalty is defined as zero, so the
first token is selected by relevance. Subsequent selections
balance task relevance and appearance diversity with
$\lambda_{\mathrm{div}}=0.25$.

Cosine similarity is computed from the raw visual features
$e_i$, before positional information is introduced. This
measures redundancy in visual appearance. The finalized
selection set defines the binary write mask
$m_i=\mathbb I[i\in\mathcal S]$. All visual tokens remain
available to the navigation backbone.

% AUTHOR CHECK: report the actual token budget K used in the experiments.

\section{Training Objectives and Test-Time Training Pipeline}
\label{app:training}

NavTTT is jointly trained for navigation with Adaptive
Parametric Memory (APM) integrated from the outset. During
inference, the learned slow-weight parameters remain fixed
and episode-local fast weights are updated online.

\subsection{Joint navigation training}

NavTTT is initialized from
Qwen3-VL-8B. The
language backbone, multimodal projector, and APM modules,
including the token scoring network and learned fast-weight
initialization, are jointly optimized under navigation
supervision. The vision encoder remains frozen. This
training adapts the policy together with memory writing
and retrieval.

Let $\theta$ denote the trainable parameters and
$\mathcal D$ the navigation training set. We minimize the
autoregressive cross-entropy over reference action tokens,
\begin{equation}
 \mathcal L_{\mathrm{nav}}(\theta)
 =-\mathbb E_{\mathcal D}\left[
 \frac{1}{N_t}\log P_\theta
 (A_t^\star\mid\mathcal X_t,M_{t-1})
 \right],
 \label{eq:app_stage1_loss}
\end{equation}
where the expectation averages over training decisions,
$A_t^\star$ is the reference action chunk, and $N_t$ is its
supervised token count. The sequence likelihood factorizes
autoregressively over the reference action tokens,
conditioned on navigation inputs $\mathcal X_t$ and the
memory state $M_{t-1}$ available before the current decision.
Prompt context and padding tokens are excluded from the loss.

% AUTHOR CHECK: describe the gradient path or estimator used to train
% the scoring network through the discrete token selection procedure.

\subsection{Self-supervised fast-weight updates}

After joint training, the slow-weight parameters are frozen.
These include the navigation backbone, multimodal projector,
token scoring network, and projections used for memory writing
and retrieval. Online adaptation updates the episode-local
fast weights $W$.

At decision $t$, the scoring network computes relevance
scores from the current visual features and navigation
context. The procedure in
Section~\ref{app:redundancy_suppression} selects tokens,
which are encoded into contextual key--value pairs. Scores
and selections vary with the current inputs.

The memory association objective is
\begin{equation}
 \mathcal L_{\mathrm{mem}}(W)
 =-\sum_i m_i\langle f_W(k_i),v_i\rangle,
 \label{eq:app_memory_loss}
\end{equation}
where $k_i$ and $v_i$ denote the encoded key and value,
$m_i$ is the selection mask, and $f_W$ is the nonlinear
fast-weight mapping. Layer and head indices are omitted.
The keys, values, and write mask are held fixed when
differentiating with respect to $W$. In the basic
gradient-descent form, the update is
\begin{equation}
 W_t=W_{t-1}-\eta_t\nabla_W
 \mathcal L_{\mathrm{mem}}(W_{t-1}),
 \label{eq:app_fast_update}
\end{equation}
where $\eta_t$ is the update rate. These updates incorporate
observed evidence without requiring ground-truth action labels.

% AUTHOR CHECK: specify the actual online optimizer, update rate,
% number of updates per decision, and any clipping or normalization.

\subsection{Causal read-before-write execution}

Prompt processing and action generation at step $t$ query
$M_{t-1}$. This state contains experience from previous
decisions and remains unchanged throughout current action
generation. Once the final decision representation is
available, the selected evidence from the current observation
produces a pending update. The updated state is committed
after the current action chunk is finalized, producing
$M_t$ for subsequent decisions.

\begin{figure*}[!t]
 \centering
 \includegraphics[width=0.85\textwidth,
  height=0.70\textheight,keepaspectratio]{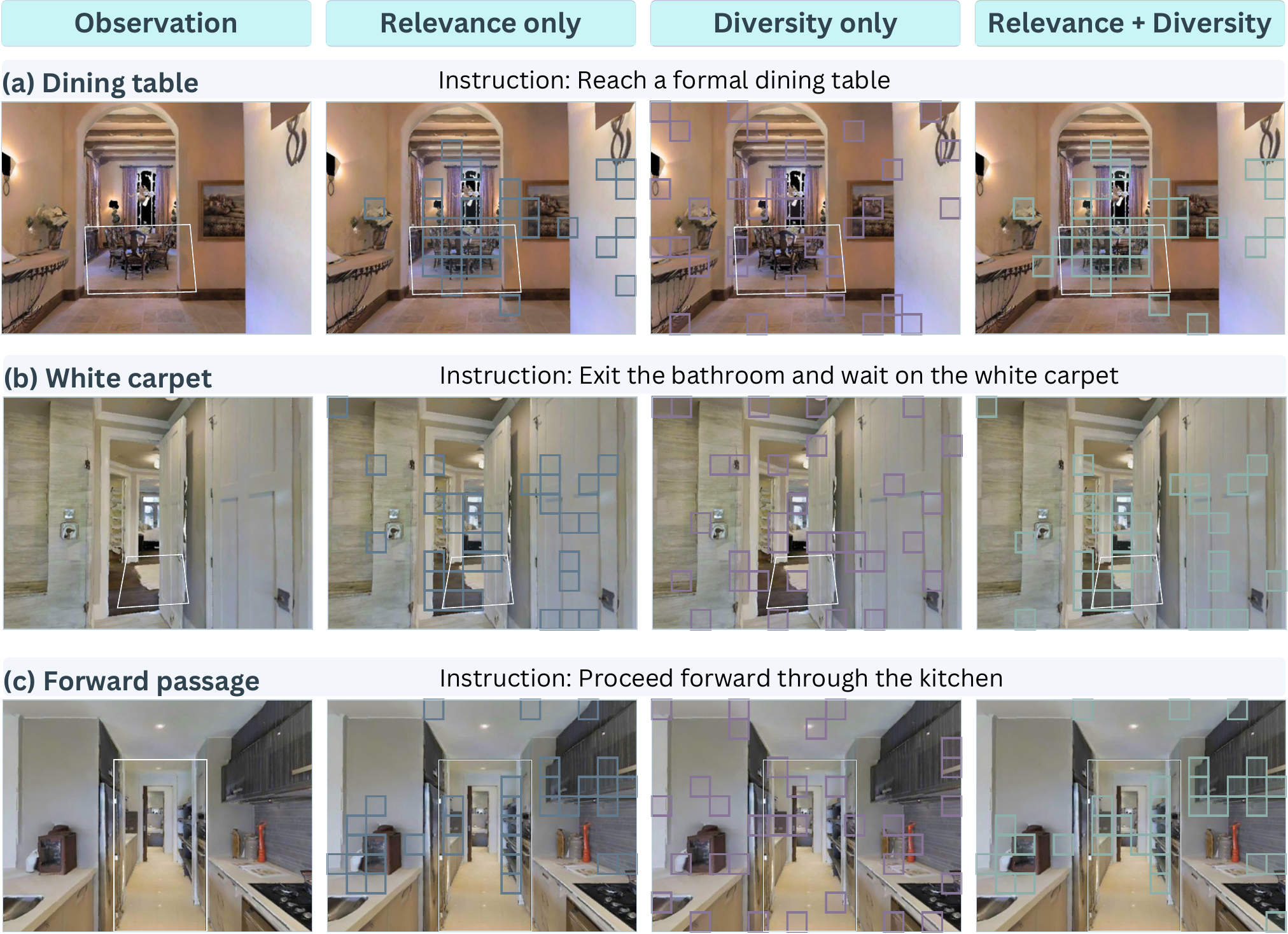}
 \caption{Qualitative comparison of relevance-only,
 diversity-only, and combined visual token selection.
 White outlines mark task-relevant regions, and colored
 boxes indicate selected visual tokens. The examples are
 screened offline comparisons under matched observations
 and decision contexts.}
 \label{fig:appendix_selector_cases}
\end{figure*}

\begin{table*}[!t]
 \centering
 \caption{Navigation performance under the same token
 budget. SR and SPL are percentages. Best scores are bold.}
 \label{tab:selection_ablation}
 \small
 \setlength{\tabcolsep}{3.5pt}
 \begin{tabular}{@{}lcccc@{}}
  \toprule
  & \multicolumn{2}{c}{R2R-CE Val-Unseen}
  & \multicolumn{2}{c}{RxR-CE Val-Unseen} \\
  \cmidrule(lr){2-3}
  \cmidrule(lr){4-5}
  Selection & SR$\uparrow$ & SPL$\uparrow$
            & SR$\uparrow$ & SPL$\uparrow$ \\
  \midrule
  Random         & 62.8 & 56.8 & 60.5 & 49.6 \\
  Relevance only & 64.1 & 58.3 & 62.7 & 51.8 \\
  Diversity only & 63.5 & 57.3 & 61.8 & 50.7 \\
  Combined & \textbf{66.7} & \textbf{59.4}
           & \textbf{63.6} & \textbf{52.7} \\
  \bottomrule
 \end{tabular}
\end{table*}

\section{Visual Token Selection Analysis}
\label{app:visual_token_selection}

We compare random, relevance-only, diversity-only, and
combined selection through full navigation rollouts under
the same evaluation settings and token budget.
Table~\ref{tab:selection_ablation} reports navigation
performance. Figure~\ref{fig:appendix_selector_cases}
illustrates the selected visual regions under matched
observations and decision contexts.

Both relevance-only and diversity-only selection improve
SR and SPL over random selection. Relevance-only selection
achieves higher scores than diversity-only selection on
both benchmarks. Combining the criteria yields the best
performance, improving SR over relevance-only selection
by 2.6 percentage points on R2R-CE and 0.9 points on
RxR-CE. These results support using task relevance to
identify useful evidence and diversity to reduce overlap
within the selected set.

The visualizations illustrate the complementary effects.
Relevance-only selection concentrates on instruction-related
content. Diversity-only selection distributes tokens more
broadly across the scene. Combined selection retains
evidence around the dining table and includes cues around
the white carpet and forward passage, with lower similarity
among the selected features. These examples illustrate
selection around compact landmarks and extended route
structures.

\begin{table*}[!t]
 \centering
 \caption{Memory configurations on R2R-CE Val-Unseen.
 H0 is the full NavTTT model. SR and SPL are percentages.
 Persistent fast-state storage is reported separately from
 total model memory. Latency is the mean per decision.
 Best navigation scores are bold.}
 \label{tab:memory_configuration}
 \small
 \setlength{\tabcolsep}{6pt}
 \begin{tabular}{@{}clrrrr@{}}
  \toprule
  ID & Configuration & SR$\uparrow$ & SPL$\uparrow$
     & State (MiB) & Latency (ms) \\
  \midrule
  H0 & Six-layer independent
     & \textbf{66.7} & 59.4 & 576.28 & 347.51 \\
  H1 & Last-layer only
     & 64.7 & 58.9 & 96.05 & 313.37 \\
  H2 & Three-layer independent
     & 65.1 & 59.2 & 288.14 & 327.81 \\
  H3 & Six-layer shared
     & 65.3 & \textbf{59.5} & 96.05 & 345.32 \\
  \bottomrule
 \end{tabular}
\end{table*}

\begin{table*}[!t]
 \centering
 \caption{Closed-loop performance by reference-route length.
 SR and SPL are percentages. Bold marks the highest score
 for each metric within a group. For full NavTTT on RxR-CE,
 SR is taken from the ablation summary and SPL from the
 benchmark record.}
 \label{tab:route_length}
 \small
 \setlength{\tabcolsep}{3pt}
 \begin{tabular}{@{}llr*{4}{rr}@{}}
  \toprule
  & & & \multicolumn{2}{c}{NavTTT (full)}
      & \multicolumn{2}{c}{Without APM}
      & \multicolumn{2}{c}{Static APM}
      & \multicolumn{2}{c}{Direct LaCT} \\
  \cmidrule(lr){4-5}
  \cmidrule(lr){6-7}
  \cmidrule(lr){8-9}
  \cmidrule(lr){10-11}
  Dataset & Group & Episodes
      & SR$\uparrow$ & SPL$\uparrow$
      & SR$\uparrow$ & SPL$\uparrow$
      & SR$\uparrow$ & SPL$\uparrow$
      & SR$\uparrow$ & SPL$\uparrow$ \\
  \midrule
  R2R-CE & Short & 615
      & \textbf{74.96} & \textbf{65.89}
      & 71.54 & 63.69 & 74.47 & 64.39 & 73.66 & 63.15 \\
         & Medium & 612
      & 71.08 & \textbf{63.32}
      & 67.65 & 60.14 & \textbf{71.57} & 62.08 & 68.63 & 61.49 \\
         & Long & 612
      & \textbf{53.92} & \textbf{48.84}
      & 47.88 & 45.74 & 43.95 & 43.89 & 47.55 & 44.53 \\
  \midrule
  RxR-CE & Short & 1,224
      & \textbf{79.33} & 70.34
      & 77.78 & 70.53 & 78.43 & \textbf{70.81} & 78.84 & 69.44 \\
         & Medium & 1,224
      & \textbf{64.95} & 51.25
      & 63.89 & \textbf{55.60} & 61.52 & 52.58 & 62.09 & 53.71 \\
         & Long & 1,221
      & \textbf{46.44} & 36.47
      & 38.82 & 32.43 & 43.98 & 34.21 & 46.03 & \textbf{37.91} \\
  \bottomrule
 \end{tabular}
\end{table*}

\section{Memory Configuration Analysis}
\label{app:memory_configuration}

We examine memory depth and state sharing on R2R-CE
Val-Unseen using Qwen3-VL-8B. All variants follow the main
joint-training protocol with matched inputs, training data,
and optimization budget. The token selector, writing feature
source, number of memory heads, and per-head state dimensions
are fixed. Online updates are enabled for every configuration.

\paragraph{Layer placement.}
We divide the 36-layer decoder into equally sized segments
and attach a memory branch to the final layer of each
segment. Using zero-based indices, the single-layer
configuration uses layer 35, the three-layer configuration
uses layers $\{11,23,35\}$, and the six-layer configuration
uses layers $\{5,11,17,23,29,35\}$. These nested layouts
retain the deepest access point while extending memory
access to earlier decoder stages.

\paragraph{State sharing.}
Independent configurations maintain a separate fast-weight
state at each augmented layer. The shared configuration
uses the same six access locations, with corresponding
memory heads sharing their states across layers. With
per-head dimensions fixed, independent configurations
allocate more total state as the number of augmented
layers increases.

Among independent configurations, increasing the number
of memory depths improves both SR and SPL, accompanied
by higher state storage and latency
(Table~\ref{tab:memory_configuration}). The six-layer shared
configuration uses the same state storage as the last-layer
configuration and improves both SR and SPL by 0.6 percentage
points. This storage-matched comparison supports accessing
historical information at multiple decoder depths.

At the same six access locations, independent states improve
SR over shared memory by 1.4 points. Shared memory achieves
0.1 point higher SPL and reduces state storage to the
single-layer footprint. Its mean latency remains close to
that of the full model. Thus, the storage saving is accompanied
by a small latency reduction. The full configuration achieves
the highest SR, and the shared configuration achieves the
highest reported SPL with lower persistent storage.

% AUTHOR CHECK: H0 latency here is 347.51 ms, while the full-model
% latency in Table S6 is 331.95 ms. Confirm both measurements and
% report their respective hardware and timing settings if they differ.

\section{Navigation Performance by Route Length}
\label{app:route_length}

We partition Val-Unseen episodes into short, medium, and
long groups using tertiles of reference trajectory length.
The cutoffs are approximately 7.8 and 10.0\,m on R2R-CE
and 9.5 and 17.7\,m on RxR-CE. The groups contain 1,839
R2R-CE episodes and 3,669 RxR-CE episodes. All variants
use the same episode groups within each benchmark.
The baseline omits APM. Static APM uses the full-model
checkpoint with online fast-weight updates disabled.

% AUTHOR CHECK: specify the RxR-CE language/subset used for these
% 3,669 episodes, and confirm that the SR and SPL records refer to
% the same checkpoint and episode list.

Full NavTTT has its largest SR advantage over the baseline
without APM in the long-route group on both benchmarks
(Table~\ref{tab:route_length}). The static control further
distinguishes the effects of memory initialization and
online updates. On R2R-CE, static APM improves SR over
the baseline on short and medium routes and scores lower
on long routes. Enabling updates increases long-route SR
by 9.97 percentage points and SPL by 4.95 points.
The short- and medium-group changes in success count
cancel, so the long group accounts for the net gain of
61 successes over static APM. The SPL gain is driven
by more successful episodes, with lower mean normalized
path efficiency among successful full-model episodes.

On long RxR-CE routes, static APM improves SR over the
baseline by 5.16 points, and full NavTTT adds 2.46 points.
The SR gain over static APM is larger in the medium group.
The comparison with direct LaCT also varies by benchmark.
Full NavTTT improves long-route SR by 6.37 points on
R2R-CE and 0.41 points on RxR-CE. Direct LaCT achieves
higher SPL on long RxR-CE routes. The results show that
the gains in task completion vary with route length and
are accompanied by different path-efficiency tradeoffs.

\section{Memory Diagnostics and Inference Profiling}
\label{app:memory_retention}

\subsection{Historical readout}

We replay 199 long RxR-CE reference routes across eight
scenes with fixed learned parameters and online fast-weight
updates. From each source write, we probe two predetermined
key--value pairs. Each probe queries the same memory head
with the original write-time key immediately after writing
and after 8, 16, or 32 further updates. We rank the same
32 values from the source write by cosine similarity to
each readout. A hit occurs when the paired value ranks first.

At each delay, we retain probes whose source frames have
left the active visual context and use the same probe set
for the immediate measurement. We average within routes,
weight routes equally, and estimate uncertainty in the
paired change using 2,000 scene-level bootstrap replicates.

\begin{table}[!t]
 \centering
 \caption{First-rank hit rates on long RxR-CE reference
 routes after subsequent memory updates. At each delay,
 source frames have left the active visual context.}
 \label{tab:historical_readout}
 \small
 \begin{tabular}{@{}lrrr@{}}
  \toprule
  Subsequent updates & 8 & 16 & 32 \\
  \midrule
  Hit rate (\%) & 6.22 & 6.05 & 5.23 \\
  \bottomrule
 \end{tabular}
\end{table}

After 32 updates, the paired change relative to the
immediate readout is $+0.14$ percentage points
(95\% CI $[-0.57,+0.73]$). The low absolute hit rate
indicates limited discrimination among values written
together. The confidence interval is consistent with small
changes in either direction under continued writing.

\subsection{Route history and action prediction}

We evaluate the same checkpoint on R2R-CE and long RxR-CE
reference routes. At each probed decision, we fix the
instruction, visual context, and pose, then encode the
prompt afresh under three fast states. The correct state
comes from the current route. Reset restores the learned
initial state. Wrong history uses a state from a different
route after the same number of updates. Replay continues
with the correct state after scoring.

We score the same reference-action tokens under teacher
forcing. For each control $c$, we compute the paired
negative log-likelihood difference
\begin{equation}
 \Delta_c=\mathrm{NLL}_c-\mathrm{NLL}_{\mathrm{correct}},
 \label{eq:app_history_nll_difference}
\end{equation}
in nats per action token. Positive values indicate higher
reference-action likelihood with correct history.
Wrong-history comparisons use decisions with a matched
donor state. Confidence intervals are obtained by resampling
scenes.

\begin{table}[!t]
 \centering
 \caption{Reference-action NLL differences in nats per
 action token. Positive values favor correct route history.
 All reported confidence intervals are above zero.}
 \label{tab:action_history_controls}
 \small
 \setlength{\tabcolsep}{2.5pt}
 \begin{tabular}{@{}lrc@{}}
  \toprule
  Control state & $\Delta_c$ & 95\% CI \\
  \midrule
  \multicolumn{3}{@{}l}{\textit{R2R-CE}} \\
  Reset & $+0.008204$ & $[0.007885,\,0.011118]$ \\
  Wrong history & $+0.005426$ & $[0.004127,\,0.011335]$ \\
  \midrule
  \multicolumn{3}{@{}l}{\textit{RxR-CE (long)}} \\
  Reset & $+0.006827$ & $[0.004219,\,0.007660]$ \\
  Wrong history & $+0.004970$ & $[0.003809,\,0.010436]$ \\
  \bottomrule
 \end{tabular}
\end{table}

The reset comparisons show a modest improvement in
reference-action likelihood from accumulated updates on
both replay sets (Table~\ref{tab:action_history_controls}).
The wrong-history comparisons show that the route supplying
those updates matters at a matched update count.
This intervention measures the contribution of the full
fast state to reference-action scoring.

\begin{table}[!t]
 \centering
 \caption{Inference cost on R2R-CE Val-Unseen. Decision
 latency includes memory writing. Peak memory is the
 maximum allocated GPU memory.}
 \label{tab:inference_efficiency}
 \small
 \setlength{\tabcolsep}{3pt}
 \begin{tabular}{@{}lrrr@{}}
  \toprule
  Variant & Decision & Writing & Peak \\
          & (ms) & (ms) & (GiB) \\
  \midrule
  Baseline (w/o APM) & 288.69 & \textemdash & 16.82 \\
  Static APM        & 318.82 & \textemdash & 17.62 \\
  NavTTT (full)     & 331.95 & 14.28       & 18.17 \\
  \bottomrule
 \end{tabular}
\end{table}

\subsection{Inference profiling}

We profile all variants on one RTX 5090 D v2 using the
same 128 R2R-CE Val-Unseen inputs, batch size 1, and four
generated tokens per decision. We average three repetitions
with 30 warm-up decisions before each measurement block.
Decision timing covers input transfer through action
detokenization and any write for the next decision. CUDA
events measure token selection, key--value preparation,
and fast-weight optimization within that interval.

Memory writing takes 14.28\,ms per decision, including
selection, key--value preparation, and fast-weight
optimization. Total decision latency increases by
13.13\,ms (4.12\%) relative to static APM
(Table~\ref{tab:inference_efficiency}).
Section~\ref{app:memory_configuration} reports latency
and persistent fast-state storage across memory layouts.
Section~\ref{app:route_length} reports the corresponding
closed-loop performance by reference-route length.

\section{More Qualitative Results}
\label{app:qualitativeresults}

We present Habitat navigation trajectories to illustrate
instruction following and representative failure modes.

\begin{figure*}[!t]
 \centering
 \includegraphics[width=0.95\textwidth,
  height=0.85\textheight,keepaspectratio]{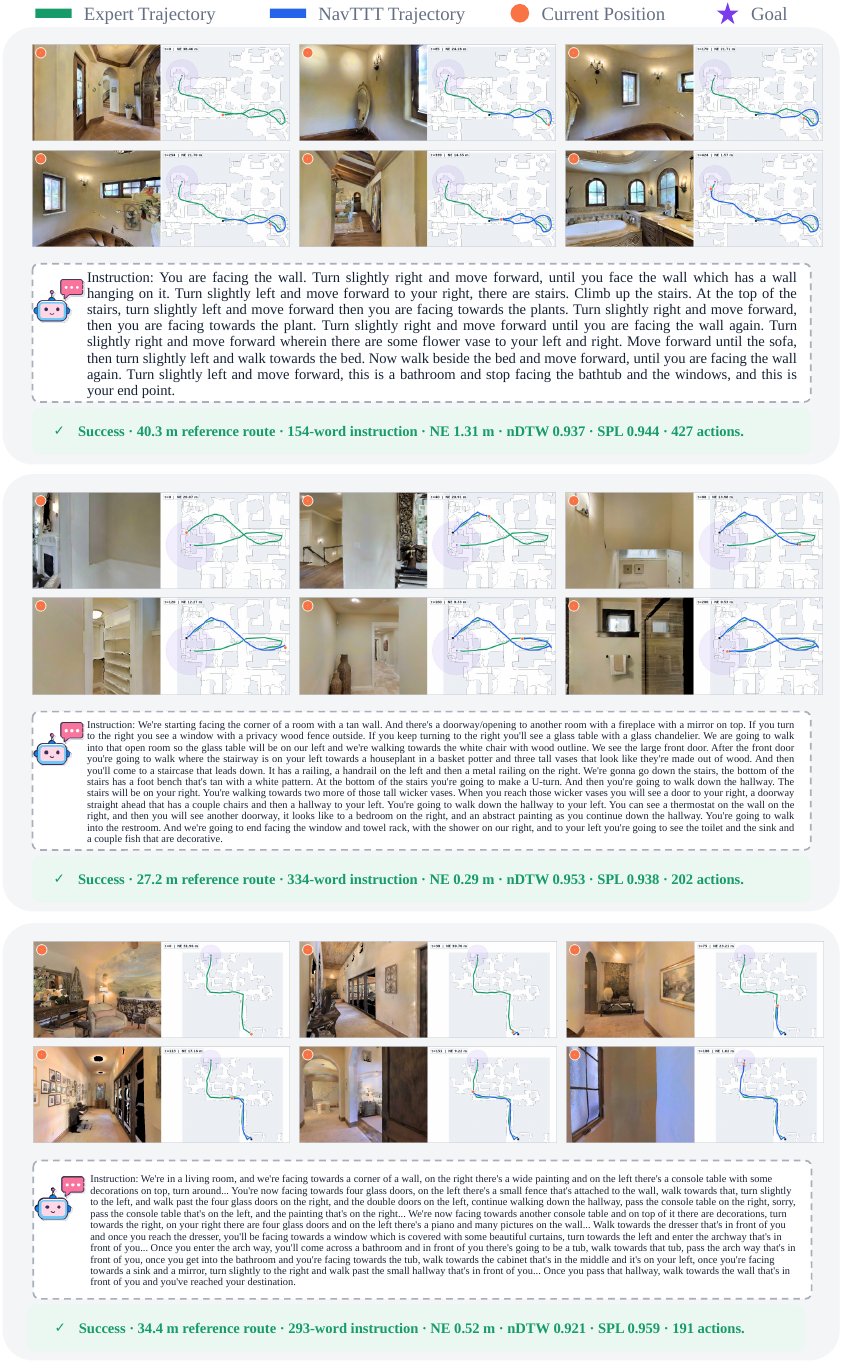}
 \caption{Qualitative navigation results of NavTTT on RxR-CE.}
 \label{fig:rxr}
\end{figure*}

\begin{figure*}[!t]
 \centering
 \includegraphics[width=0.95\textwidth,
  height=0.85\textheight,keepaspectratio]{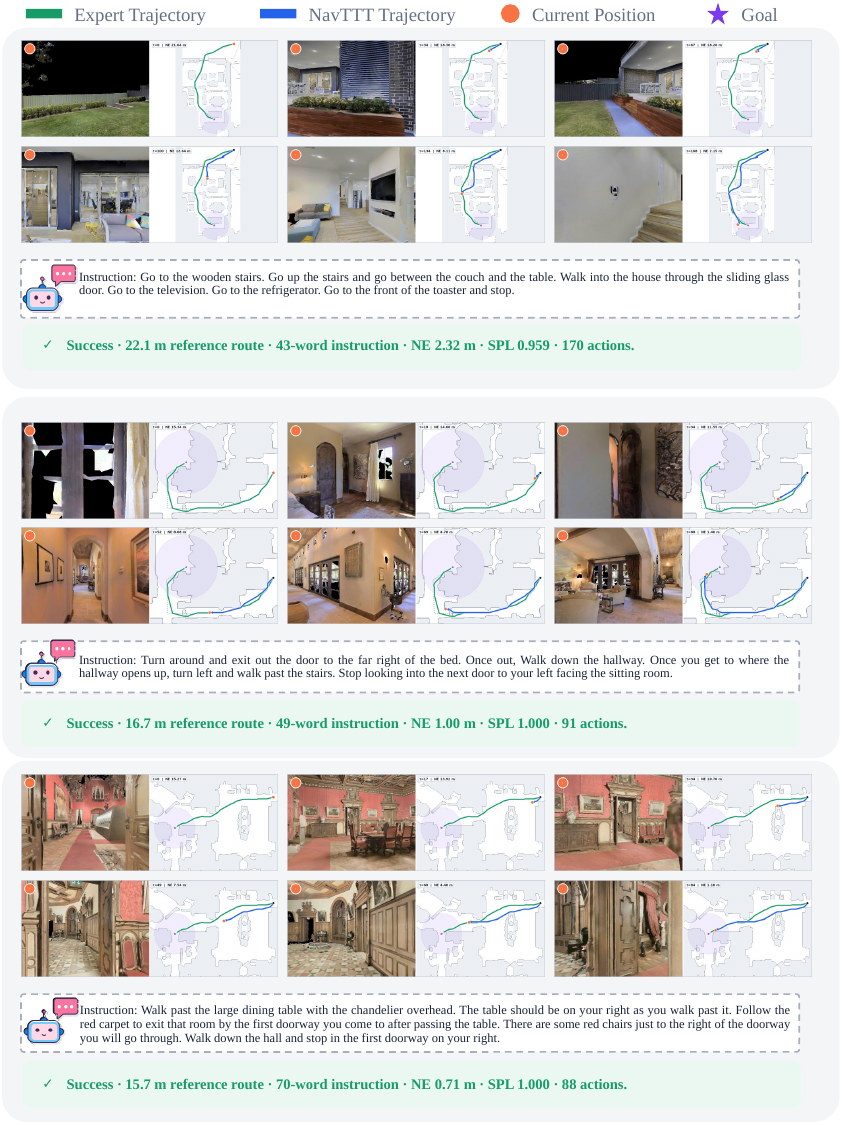}
 \caption{Qualitative navigation results of NavTTT on R2R-CE.}
 \label{fig:r2r}
\end{figure*}

\begin{figure*}[!t]
 \centering
 \includegraphics[width=0.95\textwidth,
  height=0.85\textheight,keepaspectratio]{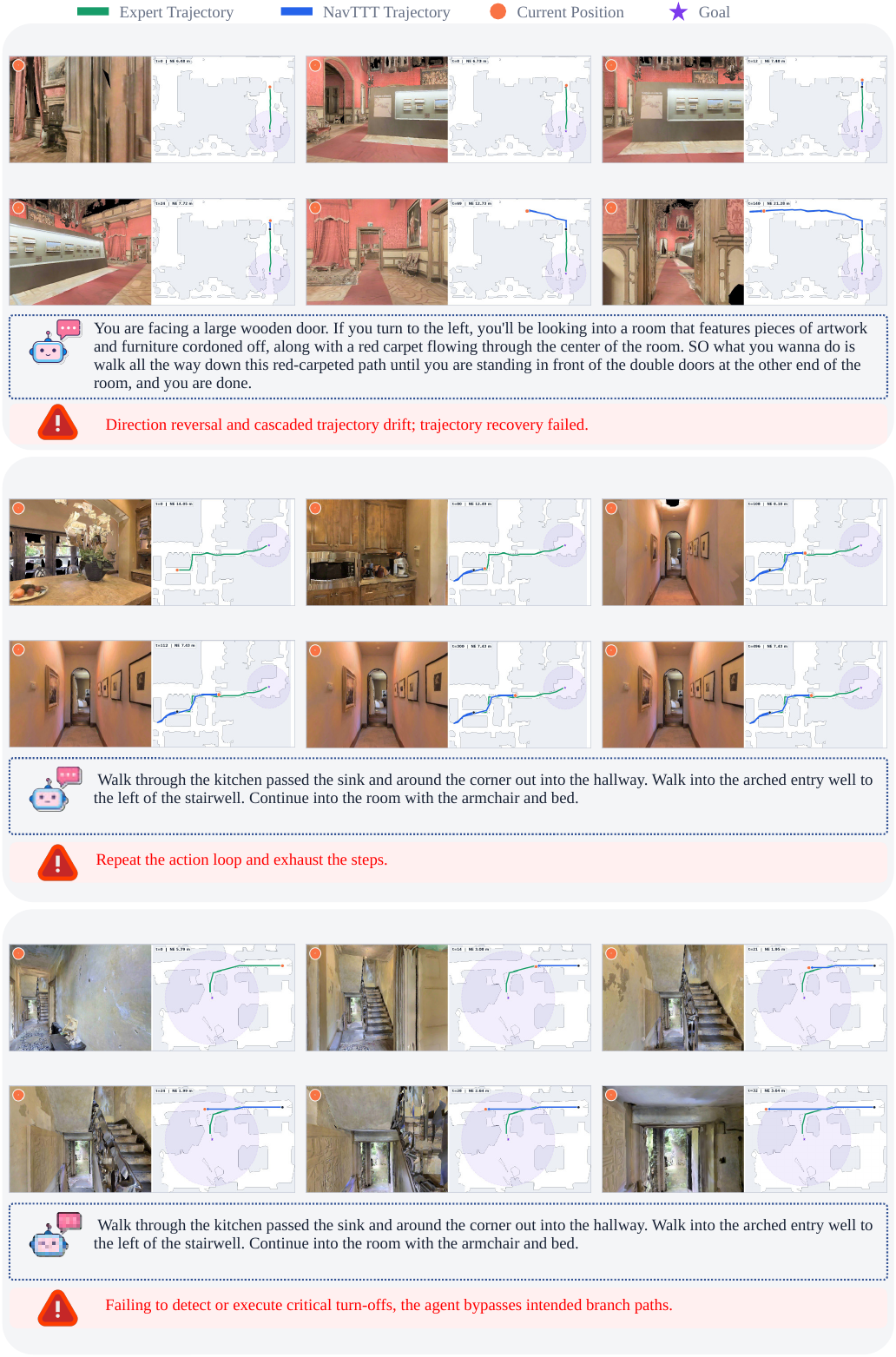}
 \caption{Representative navigation failures, including
  directional drift or reversal, repetitive action loops,
  and missed turn-offs.}
 \label{fig:failure}
\end{figure*}

\paragraph{Successful navigation.}
Figures~\ref{fig:rxr} and~\ref{fig:r2r} present successful
episodes involving multi-step instructions, extended
traversal, and target landmarks. During these episodes,
NavTTT updates its fast state with selected observations
and makes the accumulated history available to later
decisions. The examples provide qualitative illustrations
of instruction following across successive observations.

\paragraph{Failure cases.}
Figure~\ref{fig:failure} presents three failure modes.
Directional drift or reversal leads the agent away from
the instructed route. Repetitive action loops yield little
progress toward the goal. Missed turn-offs occur when the
agent passes a required branching point without executing
the instructed turn. These cases reveal challenges in
translating perceptual and historical context into reliable
local decisions. Directional grounding, progress-aware
recovery, and timely landmark-conditioned turning are
directions for further investigation.